 \documentclass[smallabstract,smallcaptions]{dccpaper}

\usepackage{epsfig}
\usepackage{amsmath}
\usepackage{amssymb}
\usepackage{color}
\usepackage{url}
\usepackage{multirow}
\usepackage[misc]{ifsym}

\newlength{\figurewidth}
\newlength{\smallfigurewidth}

\begin{document}

\title
{\large
\textbf{An End-to-End Encrypted Neural Network for Gradient Updates Transmission in Federated Learning}
}

\author{%
Hongyu Li, and Tianqi Han\\[0.5em]
{\small\begin{minipage}{\linewidth}\begin{center}
\begin{tabular}{ccc}
DL Lab, AI Institute & \hspace*{0.5in} & AI Lab\\
Tongdun Technology && ZhongAn Technology\\
Shanghai, China&& Shanghai, China\\
\Letter hongyu.li@tongdun.net&&hantianqi@zhongan.io
\end{tabular}
\end{center}\end{minipage}}
}

\maketitle
\thispagestyle{empty}

\begin{abstract}
Federated learning is a distributed learning method to train a shared model by aggregating the locally-computed gradient updates. In federated learning, bandwidth and privacy are two main concerns of gradient updates transmission. This paper\footnote{This paper is an extended version of a summary published in the Proc. of 2019 Data Compression Conference (DCC).  The 1-page summary in the DCC proceedings can be found at: https://ieeexplore.ieee.org/document/8712695.} proposes an end-to-end encrypted neural network for gradient updates transmission. This network first encodes the input gradient updates to a lower-dimension space in each client, which significantly mitigates the pressure of data communication in federated learning. The encoded gradient updates are directly recovered as a whole, i.e. the aggregated gradient updates of the trained model, in the decoding layers of the network on the server. In this way, gradient updates encrypted in each client are not only prevented from interception during communication, but also unknown to the server. Based on the encrypted neural network, a novel federated learning framework is designed in real applications. Experimental results show that the proposed network can effectively achieve two goals, privacy protection and data compression, under a little sacrifice of the model accuracy in federated learning.

\end{abstract}

\Section{Introduction}
Recently, federated learning is proposed to learn a shared model by aggregating the locally-computed updates \cite{Mcmahan2016}\cite{Shokri2015}. In federated learning, the clients compute an updated model based on their local data and post gradient updates to the server. The server then aggregates these updates (e.g. by averaging) to construct an improved global model. One significant bottleneck of federated learning is the bandwidth for gradient updates transmission. To compress the transmitted updates, the quantization method was proposed in \cite{Seide2014,Wen2017} to use low-precision values to represent the gradient updates, the authors in \cite{Strom2015,Aji2017,Kone2016} made the updates sparse through restricting them only on a small subset of parameters. However, the gradient updates may still reveal private information---local data on a client \cite{Geyer2017}, which is the other main concern in federated learning, after compression with either quantization or sparsification.

In this paper, we propose an end-to-end encrypted neural network (ENN) for compressing and encrypting dense gradient updates simultaneously. The encrypted neural network is composed of two different sub-networks: an encoding sub-network and a decoding sub-network. In the ENN-based federated learning framework, each client uses an encoding sub-network respectively to encrypt and compress the gradient updates with a lower-dimensional coding vector. Once encrypted vectors are transmitted to the server, a decoding sub-network deployed on the server is responsible to recover the aggregated updates from encrypted vectors. In this way, the original gradient updates will never appear both during transmission and after decoding. As a result, the proposed ENN can not only compress the dense gradient updates, but also protect the privacy of each client to avoid being intercepted during communication or directly exposed to the server.

The proposed ENN is trained in an end-to-end manner, where it is assumed that the overall gradient updates can be directly aggregated on the server without the explicit reconstruction of each update.
For the purpose of network training, we generate a set of training samples under a Gaussian distribution with the mean of $0$ and the standard deviation of $0.1$, assuming that gradient updates are under the same distribution.

In this work, we attempt to solve this issue of the encryption and compression of gradient updates in federated learning to some extent and our major contributions are summarized
as follows:
\begin{enumerate}
  \item An encrypted neural network is proposed for encoding and compressing the updates on each client and decoding the aggregated updates on the server, to guarantee that the updates are both secure during communication and unexposed to the server;
  \item An ENN-based federated learning framework is designed in real applications, where the encoding sub-network is flexible to extend to as many clients as desired.
\end{enumerate}

\Section{Encrypted Neural Network}


An autoencoder \cite{LIOU201484} is a type of artificial neural network to learn low-dimensional representations of data in a high-dimensional space. In the autoencoder, the input data $\mathbf{x}\in\mathbb{R}^N$ is first mapped to a lower-dimensional vector $\mathbf{y}\in\mathbb{R}^M$. Then the decoder approximately recovers the original data $\mathbf{x}$ from the low-dimensional data $\mathbf{y}$. The autoencoder learns the input data automatically and has been widely used in lossy compression applications, such as image compression \cite{krizhevsky2011} and denoising \cite{vincent2008}.

Motivated by the autoencoder, we design an encrypted neural network (ENN) to compress and encrypt dense gradient updates simultaneously in federated learning. Similar to the autoencoder, the proposed network is comprised of two sub-networks: encoder and decoder. But there are two clear differences between the ENN and the autoencoder:
\begin{enumerate}
  \item The encoder in the ENN can be distributed to multiple clients and is therefore suitable for use in federated learning,
  \item The decoder in the ENN can accept the coding vectors from any number of clients as input and directly aggregate the updates without the explicit reconstruction.
\end{enumerate}
Since the ENN can generate a decrypted result which matches the result of the operations as if they had been performed on the plaintext, it is in essence a homomorphic encryption technique \cite{Gentry:2009:FHE} that allows computation on encrypted data without exposing sensitive data. The ENN-based federated learning framework and the structure of the proposed network will be respectively introduced in more detail below.

\SubSection{Framework for federated learning}

\begin{figure}[t]
\begin{center}
\epsfig{width=5.5in,file=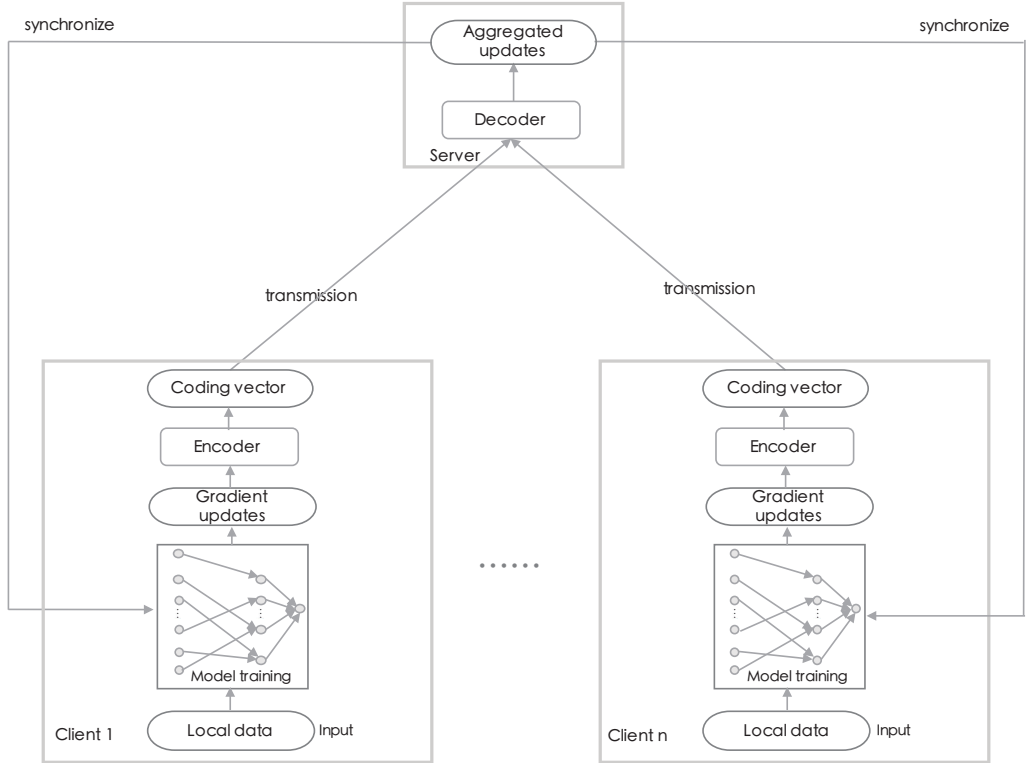} \\
\end{center}
\caption{Overview of the federated learning framework. Gradient updates for model training are first encrypted and compressed with encoder on clients before transmission and the aggregated updates are directly computed with decoder on the server.}\label{fig:fig1}%
\end{figure}

The ENN-based framework for federated learning is illustrated in Fig. \ref{fig:fig1}. In this framework, the encoder is first distributed to each client respectively and the decoder is deployed on the server.

Clients will start to train a model with local data. During training, gradient updates $\Delta \mathbf{w}$ are going to be passed to the encoder in order to encrypt and compress them. The encoder output a low-dimensional coding vector $\mathbf{y}$ and clients eventually post this vector to the server. As the encoding network is unrelated with local data on clients, the coding vector reflects no private information involving local data.

Once the server receives coding vectors from a certain number, $K$, of clients, the decoder will approximately compute the aggregated gradient updates $a(\Delta \mathbf{w}_1, \cdots, \Delta \mathbf{w}_K)$ without explicitly recovering gradient updates of any client. The server will synchronously distribute the aggregated updates to these clients, and then a new round of iteration starts for model training on clients. Since the encoder is a lossy compression, the decoder will not 100 percent reconstruct the aggregated updates. But this reconstruction way is still able to guarantee that the trained model will finally converge as expected.

\SubSection{Network structure}

Suppose the input and output data, $\mathbf{x}$ and $\mathbf{y}$, to the encoder is respectively of $N$ and $M$ dimensions. To compress gradient updates in federated learning, it is necessary to ensure $M<N$. To protect local data on clients, the updates are required to be unknown to the server. In other words, given coding vectors $(\mathbf{y}_1, \mathbf{y}_2, \cdots, \mathbf{y}_K)\in\mathbb{R}^{M\times K}$ transmitted out from $K$ different clients, the aggregated updates $a(\mathbf{x}_1, \mathbf{x}_2, \cdots, \mathbf{x}_K)\in\mathbb{R}^N$ should be directly worked out on the server. Meanwhile considering that the server usually has more powerful computing ability, we will adopt a deeper decoder but a simpler encoder in the encrypted neural network.

\begin{figure}[t]
\begin{center}
\epsfig{width=6in,file=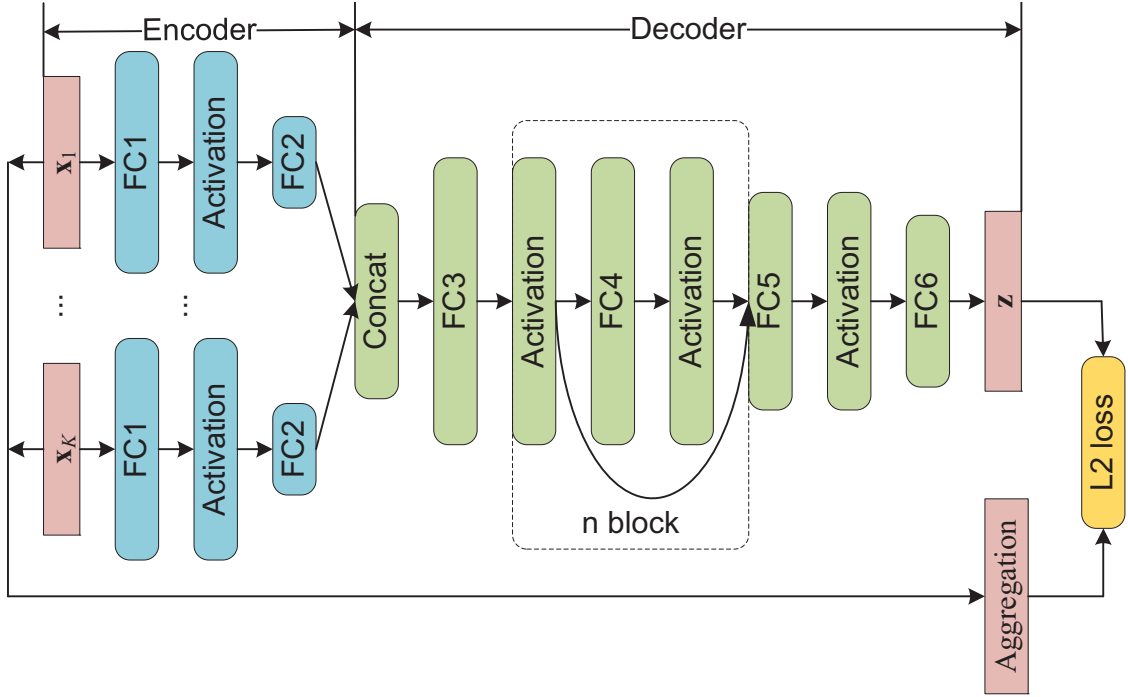} \\
\end{center}
\caption{Structure of the encrypted neural network.}\label{fig:fig2}%
\end{figure}

As shown in Fig. \ref{fig:fig2}, the whole ENN is designed to consist of multiple encoding sub-networks $f: \mathbb{R}^N\rightarrow \mathbb{R}^M$ and a decoding network $g: \mathbb{R}^{MK}\rightarrow \mathbb{R}^N$, where $K$ is the maximal number of encoders (clients) accepted by the decoder(server).

The encoding sub-network consists of $2$ fully connected (FC) layers together with an activation layer. The input data $\mathbf{x}_i\in\mathbb{R}^N$ will be mapped with the encoder to a coding vector $\mathbf{y}_i =f(\mathbf{x}_i)$ of lower dimensions $M$. Hence the compression ratio $r$ is equal to $N/M$. The related parameters are kept same for each encoding sub-network on clients.

Deeper than the encoding sub-network, the decoding one is designed with $n$ residual block \cite{He2015},where the input is the concatenation of coding vectors $\mathbf{\hat{y}} = (\mathbf{y}_1^\mathsf{T}, \cdots \mathbf{y}_K^\mathsf{T})^\mathsf{T}\in\mathbb{R}^{MK}$ and the output $\mathbf{z} =g(\mathbf{\hat{y}})$ is of $N$ dimensions. Since $\mathbf{z}$ is used to approximate the aggregation, $a(\mathbf{x}_1, \mathbf{x}_2, \cdots\mathbf{x}_K)$, of the input data $\mathbf{x}_i$, we adopt the Euclidean loss as the loss function in this network to describe the approximation. Assuming that the aggregation operation utilizes the averaging method, i.e., $a(\mathbf{x}_1, \mathbf{x}_2, \cdots\mathbf{x}_K)=\frac{1}{K}\sum_{i=1}^K {\mathbf{x}_i}$, the loss $L$ can be defined as follows,
\begin{equation*}
   L(\mathbf{x}) = \|g(\mathbf{\hat{y}}) - \frac{1}{K}\sum_{i=1}^K {\mathbf{x}_i} \|^2_,
\end{equation*}
$$s.t., \  \mathbf{\hat{y}} = (f(\mathbf{x}_1)^\mathsf{T}, f(\mathbf{x}_2)^\mathsf{T} \cdots f(\mathbf{x}_K)^\mathsf{T})^\mathsf{T}.$$

To train the proposed network, we generate a set of simulated data $\mathbf{x}$ under an assumption of the Gaussian distribution with the mean of 0 and the standard deviation of 0.1. It must be admitted that this assumption is problematic to some extent, since the distribution of gradient updates is uncertain in real applications. Fortunately, for the purpose of federated learning, the trained model will gradually converge to a certain accuracy as long as the training samples are sufficient. But the convergence speed will be slower than the normal federated learning where the ENN is not used. In addition, the model accuracy will probably drop a little due to the lossy compression caused by the encoder.
The Adam optimizer with learning rate $2e-4$ is utilized during the ENN training.

\begin{table}[!h]
\begin{center}
\caption{\label{tab:RMSE}%
Approximation error in MSE with different decoder structures and training schemes, where $n$ is the number of residual blocks in the decoding sub-network. With $n$ increasing to $3$ in the end-to-end training scheme, the error will not basically change any more.}
{
\renewcommand{\baselinestretch}{1}\footnotesize
\begin{tabular}{|c|c|c|c|c|c|}
\hline
~ &freezing decoder& freezing encoder & \multicolumn{3}{|c|}{end-to-end} \\
\cline{1-6}
 n	 & 3 &3 & 0 &3 &7 \\
\hline
MSE &0.0112  &0.00232 &0.00210  &0.00166  &0.00168\\
\hline
\end{tabular}}
\end{center}
\end{table}

\Section{Experimental Results}
In this section, we conducted two experiments to validate the performance of the proposed network. The first experiment is involved with the approximation ability of the ENN to the true aggregation of input data. In the second test, the model accuracy is analyzed on the MNIST dataset in the ENN-based federated learning framework.

\SubSection{Approximation error}
With the generated training data, we estimated the approximation errors under three different training schemes, i) training encoder only while freezing decoder with random weights, ii) training decoder only while freezing encoder with random weights, and iii) end-to-end training. In addition, we also compared three network structures of the decoder in the end-to-end training, where the number $n$ of residual blocks is respectively 0, 3, and 7.

The mean square error (MSE) was used to measure the approximation error. As shown in Table \ref{tab:RMSE}, it is observed that the end-to-end training produced lower approximation errors than solely training either the encoder or decoder. Moderately increasing the depth of the decoding sub-network can decrease the approximation error. However, the MSE becomes stable and will not change any more when the number $n$ is larger than $3$. The approximation error mainly stems from the lossy compression during encoding, which also indicates that the aggregation of the input data is hardly to be exactly recovered due to data encryption. Luckily, the main concern in the federated learning is the model accuracy that will not be affected by the slight approximation error.

\begin{figure}[!h]
\begin{center}
\epsfig{width=5in,file=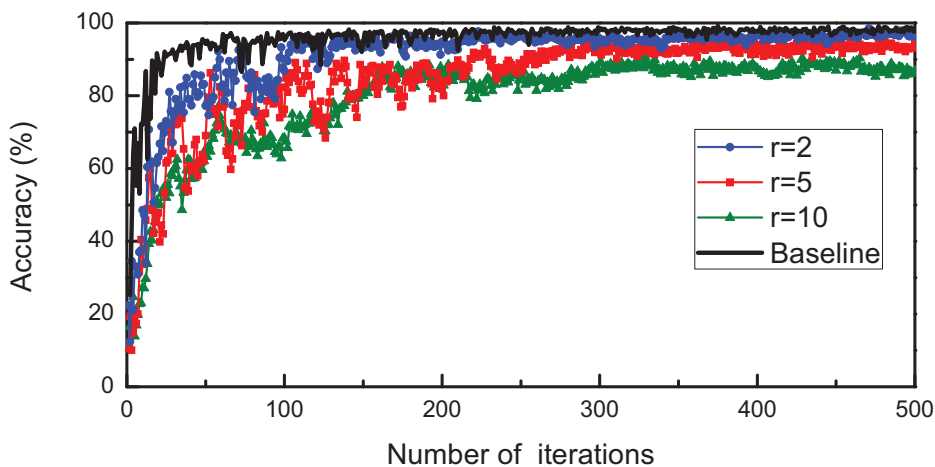} \\
\end{center}
\caption{\label{fig:fig3}%
Model accuracy under different compression ratio.}
\end{figure}

\SubSection{Accuracy analysis in federated learning}

To estimate the ENN-based federated learning framework, we simulated 100 clients connecting with a server and trained a convolutional neural network model for classification on the MNIST dataset with the proposed framework. In each iteration round, 9 clients were randomly picked from all 100 clients to train the model and update the gradients. For simplicity, a small neural network with $2$ convolutional layers was constructed with total $21840$ weights. The dataset was split into different groups as local data for model training on clients, where each group has only three types of characters. But in the test set, all types of character images were pooled together. In this way of data splitting, gradient updates of each client are indispensable to the aggregated model. For normal federated learning without using the ENN, the model accuracy can be as high as $98\%$ on the test set, which will act as a baseline in this experiment. When the ENN was applied to federated learning, weight updates will be compressed and the private local data on clients are more secure at the cost of sacrificing a little accuracy and training speed.

Fig. \ref{fig:fig3} illustrates the test accuracy under different compression ratio $r$. It is observed that the accuracy at the $2\times$ compression ratio can almost reach the baseline after about $100$ iterations. When $r=5$, the model converged more slowly and became stable after $300$ iterations at the accuracy of $93.7\%$. The larger the compression ratio, the more the test accuracy dropped. The accuracy will reduce to $87.5\%$ in the case of the $10\times$ compression ratio after $300$ iterations. In summary, the compression ratio of $5\times$ seems acceptable to balance the model accuracy and compression for gradient updates transmission.

\Section{Conclusion}
In this paper, we propose an end-to-end encrypted neural network (ENN) for compressing and encrypting dense gradient updates simultaneously. The encrypted neural network is composed of two different sub-networks: an encoding sub-network and a decoding sub-network. In the ENN-based federated learning framework, each client uses an encoding sub-network respectively to encrypt and compress the gradient updates with a lower-dimensional coding vector. Once encrypted vectors are transmitted to the server, a decoding sub-network deployed on the server is responsible to recover the aggregated updates from encrypted vectors. In this way, the original gradient updates will never appear both during transmission and after decoding. As a result, the proposed ENN can not only compress the dense gradient updates, but also protect the privacy of each client to avoid being intercepted during communication or directly exposed to the server.

\Section{References}
\bibliographystyle{IEEEbib}
\bibliography{refs}

\end{document}